\documentclass[10pt, a4paper]{article}

\usepackage[]{lrec-coling2024} 

\usepackage{amsfonts}
\usepackage{amsmath}
\usepackage{booktabs}
\usepackage{graphicx}
\usepackage{caption}
\usepackage{subcaption}
\usepackage{enumitem}

\usepackage{xcolor,pifont}
\newcommand*\colourcheck[1]{%
  \expandafter\newcommand\csname #1check\endcsname{\textcolor{#1}{\ding{52}}}%
}
\colourcheck{blue}
\colourcheck{green}
\colourcheck{red}
 
\newcommand{\xmark}{\ding{55}}
\newcommand{\redxmark}{{\color{red}\xmark}}

\newcommand{\ours}{\texttt{HumanEval-XL}}

\title{{\ours}: A Multilingual Code Generation Benchmark for Cross-lingual Natural Language Generalization}

\name{Qiwei Peng$\thanks{*Equal contribution and shared co-first authorship.}^{*\heartsuit}$, Yekun Chai$^{*\spadesuit}$, Xuhong Li$^{\spadesuit}$} 

\address{$^{\heartsuit}$University of Copenhagen \qquad $^{\spadesuit}$Baidu Inc.\\
         qipe@di.ku.dk \quad  \{chaiyekun,lixuhong\}@baidu.com\\
        }

\abstract{
Large language models (LLMs) have made significant progress in generating codes from textual prompts. However, existing benchmarks have mainly concentrated on translating English prompts to multilingual codes or have been constrained to very limited natural languages (NLs). These benchmarks have overlooked the vast landscape of massively multilingual NL to multilingual code, leaving a critical gap in the evaluation of multilingual LLMs. In response, we introduce {\ours}, a massively multilingual code generation benchmark specifically crafted to address this deficiency. {\ours} establishes connections between 23 NLs and 12 programming languages (PLs), and comprises of a collection of 22,080 prompts with an average of 8.33 test cases. By ensuring \textit{parallel} data across multiple NLs and PLs, {\ours} offers a comprehensive evaluation platform for multilingual LLMs, allowing the assessment of the understanding of different NLs. Our work serves as a pioneering step towards filling the void in evaluating NL generalization in the area of multilingual code generation. We make our evaluation code and data publicly available at \url{https://github.com/FloatAI/humaneval-xl}.
 \\ \newline  }

\begin{document}

\maketitleabstract

\section{Introduction}
Large language models (LLMs) have made significant progress in code generation, with a predominant focus on the evaluation of generating Python code within an English-centric context~\citep{chen2021codex, austin2021program, hendrycksapps2021, nijkamp2022codegen}. In recent years, there has been a growing interest in expanding beyond English-Python code generation to English-multilingual programming languages (PLs)~\citep{mxeval23}. Some works have also made efforts to incorporate different natural languages (NLs), such as Spanish and Russian, although their coverage of different natural languages remains rather limited~\citep{wang2022execution, Chai2022ERNIECodeBE, Scao2022BLOOMA1}. A noticeable gap in these benchmarks is their limited ability to provide a rigorous assessment of multilingual NL to code generation. This overlooks the wide range of different NLs, raising important questions to the evaluation of cross-lingual NL generalization of LLMs. It further underscores the need for a more comprehensive evaluation framework that addresses the gap and captures the diverse linguistic landscape in code generation.

In light of this significant gap, this work presents a massively multilingual benchmark named {\ours}. This benchmark effectively tackles previous limitations by incorporating parallel coding problems across 23 NLs and 12 PLs. This facilitates a thorough evaluation of multilingual NL to code generation. The parallel alignment of datasets across multilingual languages ensures a comprehensive assessment of cross-lingual generalization, setting a new standard for evaluating LLMs on code generation.


Moreover, we perform extensive experiments to assess the multilingual code generation capability of three distinct families of LLMs: CodeT5+, CodeGen2, GPT-3.5, and GPT-4, at various parameter scales. Our findings reveal a notable trend that substantial increases in model size significantly boost the proficiency of generating code in multiple languages, leading to a deeper understanding of both NLs and PLs. We further demonstrate that specialized code pre-training plays a pivotal role in code generation. Intriguingly, the evaluation on 23 NLs highlights a significant challenge for current LLMs, as they struggle to capture the equivalent semantic meaning expressed in different languages in the task of code generation.


\section{Related work}
\label{sec:rw}
\paragraph{Code Generation Benchmarks} Prior efforts have predominantly concentrated on English-centric Python generation, evident in benchmarks such as HumanEval~\citep{chen2021codex}, MBPP~\citep{austin2021program}, APPS~\citep{hendrycksapps2021}, DSP~\citep{chandel2022training}, MTPB~\citep{nijkamp2022codegen}, and DS-1000 \citep{lai2023ds}. Some endeavors have extended the HumanEval benchmark to multiple PLs \citep{mxeval23, 10.1145/3580305.3599790}. Conversely, ODEX~\citep{wang2022execution} introduced a dataset comprising 4 NLs and Python, albeit with a limited average of 1.8 test cases. In contrast, our work pioneers a multilingual code generation benchmark, connecting 23 NLs and 12 PLs in parallel. This extensive dataset comprises 22,080 prompts with an average of 8.33 test cases, setting a new standard for comprehensive evaluation in multilingual code generation.

\paragraph{Multilingual Code LLMs} 
LLMs undergo a critical phase of pre-training on a rich collection of resources that include both multilingual PL data and NL data. Some studies have delved into the intricacies of LLMs in multiple PLs~\citep{incoder22,wang2023codet5,nijkamp2023codegen2,starcoder23}, while others have focused on the synergy between multilingual NLs and PLs. Notably, ERNIE-Code~\citep{Chai2022ERNIECodeBE} pioneered multilingual LLMs by pre-training across 116 NLs and 6 PLs. Likewise, BLOOM~\citep{Scao2022BLOOMA1} was pre-trained on data in 46 NLs and 13 PLs. However, it did not produce competitive results in Python code generation. Recent LLMs like GPT3.5~\citep{chatgptopenai} and GPT-4~\citep{OpenAI2023GPT4TR} also cater to different NLs and PLs. A significant gap in current code generation benchmarks is the absence of multilingual code generation, preventing an assessment of the capability of LLMs on cross-lingual natural language generalization. Our benchmark aims to address this gap comprehensively.


\section{\ours}


\subsection{Design Principles}
In crafting {\ours}, our primary objective is to create an accessible code generation benchmark tailored for evaluating cross-lingual NL transfer learning across different PLs. The languages and evaluation are devised in adherence to the following guiding principles:

\paragraph{(1) Task Complexity.}
A fundamental consideration in our selection process is the complexity of the tasks at hand. We opt for code generation tasks due to their inherent challenge, providing a robust evaluation ground. To assess the code generation proficiency of LLMs, we employ the pass@k metric~\citep{chen2021codex}, gauging the success rate of executed test cases. This metric offers a nuanced understanding of the LLMs' capabilities in tackling coding challenges.

\paragraph{(2) Language Diversity.}
It is necessary to ensure multilingual diversity to develop a comprehensive evaluation platform. We therefore aim to incorporate a wide range of NLs and PLs. Moreover, to facilitate unbiased comparisons, the cases within distinct NLs are made parallel. This approach ensures a fair and insightful comparison of LLMs' performance across diverse linguistic landscapes.

\paragraph{(3) Accessibility.}
Selected data in our benchmark are sourced from materials available under permissive licenses, allowing unrestricted use and redistribution of data for research purposes.  

\subsection{Dataset Construction}
\begin{figure*}
\centering
\includegraphics[width=0.8\textwidth]{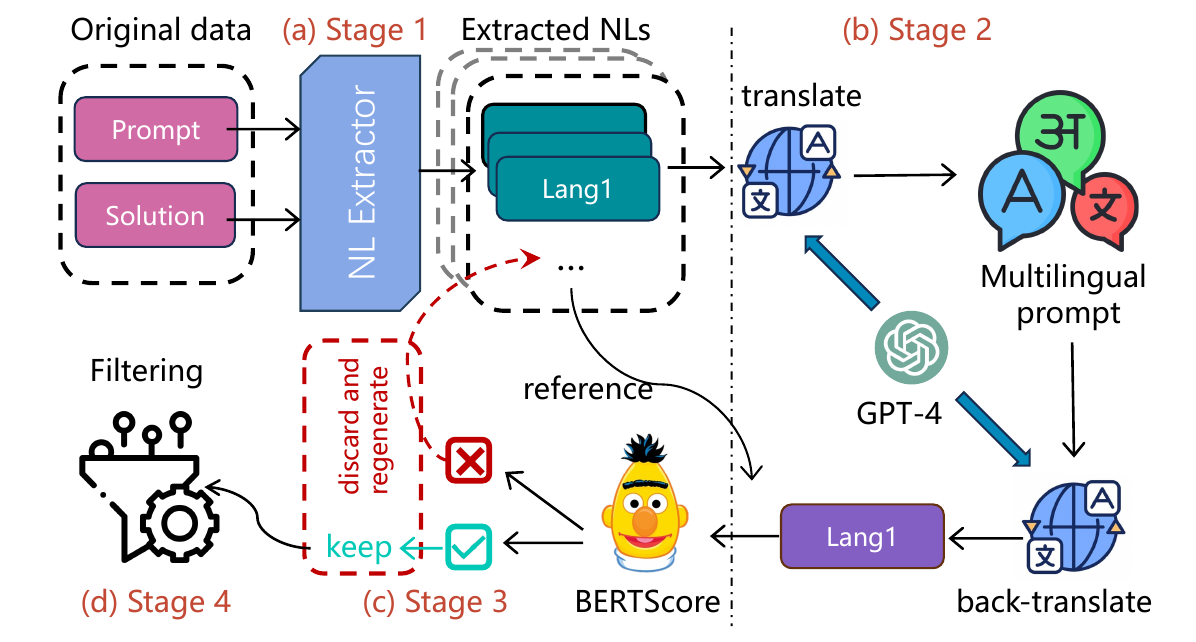}
\caption{Illustration of data construction in four steps: \textbf{(a) Text Extraction  (Stage 1)}:  We extract NL texts from the prompt in stage 1. \textbf{(b) Translation and Back-Translation  (Stage 2)}: The extracted texts are translated into 23 different languages using \textit{GPT-4}. These translations are then back-translated to English for subsequent automatic quality checks. \textbf{(c) Quality Assessment with \textit{BERTScore} (Stage 3)}: Stage 3 assesses translation quality by computing the BERTScore similarity score between the original English text and its back-translated text. Translations with a low similarity score (threshold $< 0.95$) are rejected and subjected to re-translation. \textbf{(d) Quality Control (Stage 4)}: Heuristic checks and manual evaluations are performed on the quality of the translated texts.}
\label{fig:construction-process}
\end{figure*}  

In this study, we introduce a robust and efficient method for constructing a comprehensive evaluation benchmark in the field of multilingual code generation. Our approach focuses on extending a given NL-PL pair, such as English-Python, to a diverse set of languages using an iterative paradigm by leveraging LLMs like GPT-4. As illustrated in Figure \ref{fig:construction-process}, we present a four-stage data construction process.

\paragraph{(1) Extraction and Wrapping of NL Prompt:}
Given an English prompt of an example problem, we begin by extracting the NL portion of the prompt. This extracted NL text is then wrapped into a format suitable for translation in the subsequent stages.

\paragraph{(2) Translation and Back-Translation:}
We utilize the powerful GPT-4 to translate the extracted NL text into 23 different NLs. This step ensures the diversity of the benchmark across various linguistic contexts. Simultaneously, these 23 translations are back-translated into English using GPT-4, where they will be used for a quality check in later stages.
\paragraph{(3) Quality Check with BERTScore:}
To assess the quality of the translation, we employ BERTScore \citep{zhang2019bertscore} to evaluate the similarity between the back-translated English prompt and the original English text. If the similarity score exceeds 0.95, indicating a high-quality translation, it is retained for further processing. In cases where the score falls below this threshold, above process is iteratively repeated up to a predefined limit (set to 3 in this work). If the score is still below the threshold, we will discard this example. 
\paragraph{(4) Quality Control:} Subsequently, the remaining translated texts undergo heuristic quality checks. We have manually checked randomly sampled examples and make corrections if the translation is problematic. Starting from multilingual HumanEval \citep{mxeval23}, we apply above pipeline to all examples in it and translate them into 23 different languages. After filtering and discarding poor-quality examples, we result in the \ours {} benchmark, consisting of 80 parallel coding problems spanning 12 PLs and 23 NLs. In total, this benchmark includes 22,080 coding problems. 

\begin{table}[]
\resizebox{\columnwidth}{!}{%
\begin{tabular}{@{}ll@{}}
\toprule
\textbf{Family} & \textbf{Languages} \\ \midrule
Afro-Asiatic & Arabic, Hebrew \\
Austro-Asiatic & Vietnamese \\
Austronesian & Indonesian, Malay, Tagalog \\
Indo-European (Germanic) & English, Dutch, German, Afrikaans \\
Indo-European (Romance) & Portuguese, Spanish, French, Italian \\
Indo-European (Greek) & Greek \\
Indo-European (Iranian) & Persian \\
Slavic & Russian, Bulgarian \\
Sino-Tibetan & Chinese \\
Turkic & Turkish \\
Uralic & Estonian, Finnish, Hungarian \\ \bottomrule
\end{tabular}%
}
\caption{Language families of our dataset, encompassing 11 language families and 23 languages.}
\label{tab:lang_family}
\end{table}

\paragraph{PLs and NLs} In this dataset, we consider 12 PLs and 23 NLs. The 12 PLs are the same as in Multilingual HumanEval, including Python, Java, Go, Kotlin, PHP, Ruby, Scala, JavaScript, C\#, Perl, Swift and TypeScript. 23 distinct NLs, as illustrated in Table \ref{tab:lang_family} and Table \ref{tab:data}, represent a diverse linguistic background. Notably, these languages are distributed across 11 language families, further demonstrating the dataset's comprehensive coverage and multilingual nature.



\begin{table}[th]
\resizebox{\columnwidth}{!}{%
\begin{tabular}{@{}c|cclccc@{}}
\toprule
\textbf{Dataset} & \textbf{\#Samples} & \textbf{\begin{tabular}[c]{@{}l@{}}\#Average \\ Test Cases\end{tabular}} & \textbf{Data source} & \textbf{\#PL} & \textbf{\#NL} & \textbf{Parallel?} \\ \midrule
HumanEval \citep{chen2021codex} & 164 & 7.7 & Hand-written & 1 & 1 & \redxmark \\
MBPP \citep{austin2021program} & 974 & 3.0 & Hand-written & 1 & 1 &  \redxmark\\
APPS \citep{hendrycksapps2021} & 10,000 & 13.2 & Competitions & 1 & 1 &  \redxmark\\
DSP \citep{chandel2022training} & 1,119 & 2.1 & \begin{tabular}[c]{@{}l@{}}Github \\ Notebooks\end{tabular} & 1 & 1 & \redxmark \\
MTPB \citep{nijkamp2022codegen} & 115 & 5.0 & Hand-written & 1 & 1 & \redxmark \\
DS-1000 \citep{lai2023ds} & 1,000 & 1.6 & StackOverflow & 1 & 1 & \redxmark \\
\begin{tabular}[c]{@{}l@{}}Multlingual HumanEval\\ \citep{mxeval23}\end{tabular}
 & 1,935 & 7.8 & Hand-written & 12 & 1 & \redxmark \\
ODEX \citep{wang2022execution} & 945 & 1.8 & StackOverflow & 1 & 4 & \redxmark \\ \hline
{\textbf{\ours}} & \textbf{22,080} & 8.3 & Hand-written & \textbf{12} & \textbf{23} &  \greencheck  \\ \bottomrule
\end{tabular}%
}
\caption{Comparison of {\ours} with previous code generation benchmarks in terms of sample numbers, average test cases, data source, number of supported PLs, number of supported NLs, and parallel support across different NLs.}
\label{tab:data}
\end{table}

\section{Experiments}
\subsection{Experimental Setup}
We evaluate the performance of three different LLMs on our benchmark: CodeT5+ \citep{wang2023codet5}, CodeGen2 \citep{nijkamp2023codegen2}, two GPT variants, namely GPT-3.5 \citep{chatgptopenai} and GPT-4 \citep{OpenAI2023GPT4TR}. CodeT5+ and CodeGen2 are specialized models designed for code generation tasks, whereas GPT-3.5 and GPT-4 are general-purpose LLMs. For CodeT5+, we experiment with different parameter scales including 220M, 770M and 2B. Similarly, we also evaluate CodeGen2 with 1B, 3.7B, 7B and 16B\footnote{Due to page limits, we put all experimental results in appendix and only report different models with largest parameters in the main context, as the observed trends remain consistent across different scales for each model.}. Due to constrained computing resources, we report pass@1 for all experiments.

\begin{figure*}
    \centering
    \includegraphics[width=\textwidth]{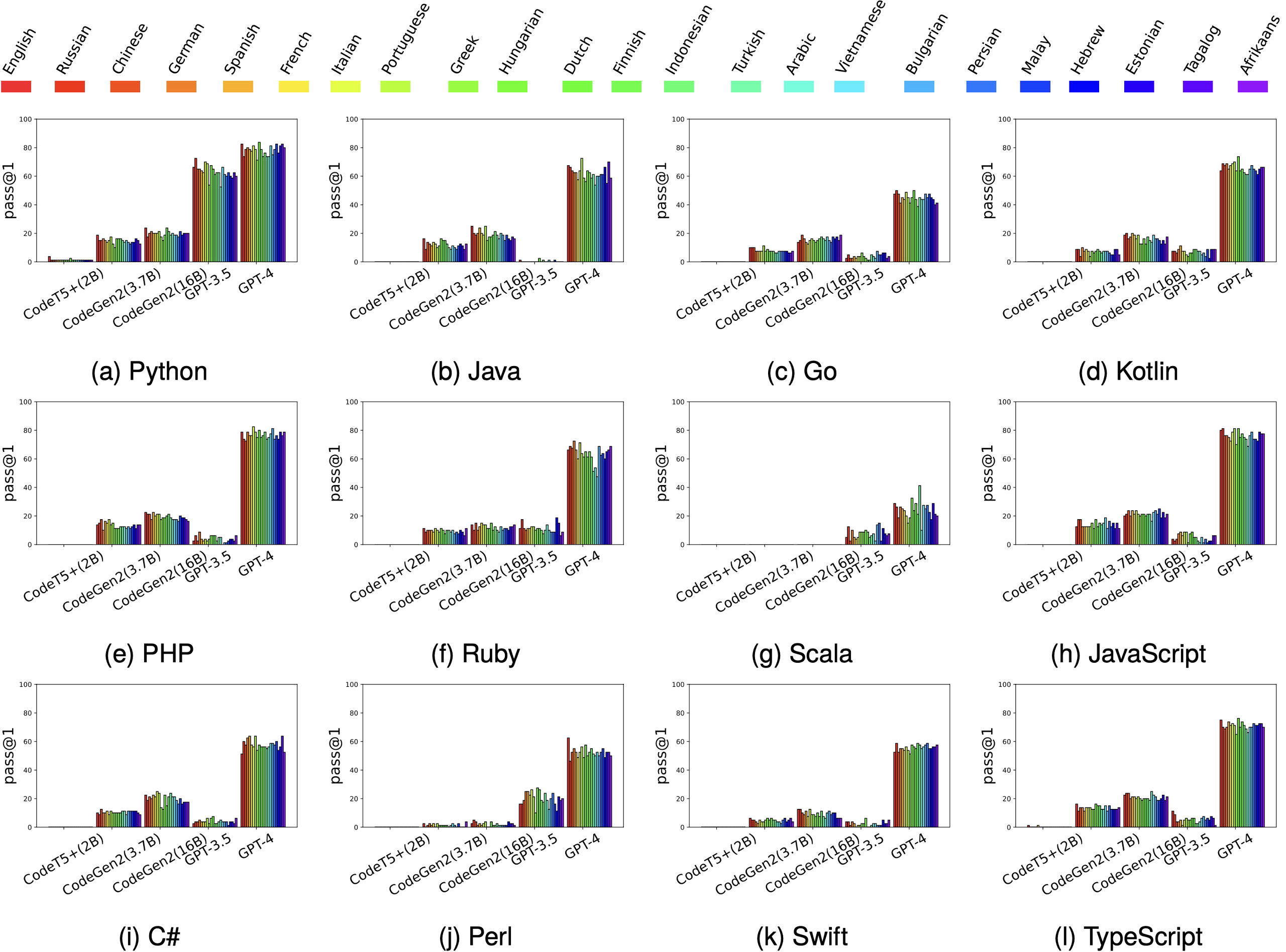}
    \caption{Performance of LLMs, including CodeT5+ (2B), CodeGen2 (3.7B, 16B), GPT-3.5, and GPT-4, on our proposed benchmark. We report pass@1 metric for evaluation. We order languages in their resource availability as summarized in CC100 XL corpus \citep{lin2022few}.}
    \label{fig:experiment_result}
\end{figure*}

\subsection{Results}
The experimental results are depicted in Figure \ref{fig:experiment_result}. We compare CodeT5+ (2B) and CodeGen2 (3.7B, 16B) against GPT-3.5 and GPT-4 across 12 different PLs. The full and detailed results for varied sizes are presented in the appendix. Each sub-figure presents a summary of the pass@1 results across 23 NLs for a fixed PL. 

From Figure \ref{fig:experiment_result}, we can observe clear disparities in performance across different NLs. This variation underscores the ongoing challenge of achieving cross-lingual NL generalization for current LLMs. Notably, GPT-4 consistently demonstrates superior performance across different PLs and NLs (except C\#), outperforming all other models significantly. Interestingly, despite the number of parameters of GPT-3.5 is considerably higher than CodeGen2-16B, a specifically tailored code generation modelLLM, GPT-3.5 lags behind CodeGen2-16B in all PLs except Python. This underscores the crucial role of training on coding data. 

CodeT5+ (2B), despite its specialization in code generation, struggles to solve problems across most scenarios. In contrast, both CodeGen2-3.7B and CodeGen2-16B exhibit markedly improved performance compared to CodeT5+ (2B). This divergence in performance is likely attributed to the difference in model structure. CodeT5+ follows an encoder-decoder architecture, while all other models employ a decoder-only structure. Our findings suggest that without further instruction-tuning, the encoder-decoder structure of CodeT5+ hinders its ability to complete code generation tasks proficiently across various PLs and NLs.

Python stands out as the most effectively tackled PL by evaluated LLMs. Conversely, all models seem to struggle with Scala and Go, as depicted in the figure. This highlights the formidable complexity of code generation tasks in a multilingual PL context.

\subsection{Analysis and Discussion}
To further investigate the disparity in performance across different NLs, we have initially categorized the 23 NLs into three distinct groups based on resource availability, following the taxonomy outlined in \citet{joshi-etal-2020-state} (ranging from 0 = least resourced to 5 = best resourced). Class 5 contains EN, ES, FR, ZH, AR, DE. Class 4 contains PT, IT, NL, RU, FI, VI, HU, FA. Class 3 contains AF, ID, BG, EL, TL, MS, HE, ET, TR. Specifically, we examine the performance of different models on Python across grouped NLs. Results are presented in Table \ref{tab:low_high_result}. For brevity, full results will be provided in the appendix due to space constraints. 


\begin{table}[h!]
\centering
\resizebox{0.8\columnwidth}{!}{%
\begin{tabular}{@{}l|ccc@{}}
\toprule
\textbf{Model}           & \textbf{Class 5} & \textbf{Class 4} & \textbf{Class 3} \\ \midrule
CodeT5+ (2B)    & 0.63{\small $\pm$1.53}             & 0.94{\small $\pm$0.88}             & 0.83{\small $\pm$0.63}             \\
CodeGen2 (3.7B) & 15.42{\small $\pm$1.88}            & 14.69{\small $\pm$2.39}            & 14.31{\small $\pm$1.41}            \\
CodeGen2 (16B)  & 20.83{\small $\pm$1.51}            & 19.06{\small $\pm$2.65}            & 19.58{\small $\pm$1.25}            \\
GPT-3.5         & 62.50{\small $\pm$5.06}            & 66.41{\small $\pm$4.25}            & 60.42{\small $\pm$2.86}            \\
GPT-4           & \textbf{78.54{\small $\pm$2.90}}            & \textbf{78.75{\small $\pm$3.54}}            & \textbf{77.64{\small $\pm$4.07}}            \\ \bottomrule
\end{tabular}%
}
\caption{Performance of different models on Python across grouped NLs, categorized according to the taxonomy of \citet{joshi-etal-2020-state} in resource availability. Pass@1 scores averaged across different languages within each group are reported along their standard deviation.}
\label{tab:low_high_result}
\end{table}

Interestingly, we observe from Table~\ref{tab:low_high_result} that LLMs within the same family, but possessing different parameter scales, exhibit a high correlation across different levels of language resources. For instance, the Pearson correlation between CodeGen2-3.7B and 16B is 0.8, and between GPT-3.5 and GPT-4, it stands at 0.87. This finding underscores the persistence of scaling laws~\cite{chinchilla22} while maintaining a consistent landscape of capabilities across a wide range of linguistic structures. However, LLMs from different families have no such correlation. Our research sheds light on the intriguing stability in performance observed within LLMs, regardless of their varying parameter scales.

\begin{table}[h!]
\centering
\resizebox{0.9\columnwidth}{!}{%
\begin{tabular}{@{}l|ccccc@{}}
\toprule
\textbf{Language Family}           & \textbf{CodeT5+ (2B)} & \textbf{CodeGen2 (3.7B)} & \textbf{CodeGen2 (16B)} & \textbf{GPT-3.5} & \textbf{GPT-4} \\ \midrule
Afro-Asiatic             & 0.63{\small $\pm$0.88}                  & 13.75{\small $\pm$0.00}                    & 19.38{\small $\pm$0.88}                   & 56.25{\small $\pm$5.30}   & 75.00{\small $\pm$1.77} \\
Austro-Asiatic           & 1.25{\small $\pm$0.00}                  & 15.00{\small $\pm$0.00}                    & 18.75{\small $\pm$0.00}                   & 66.25{\small $\pm$0.00}   & 81.25{\small $\pm$0.00} \\
Austronesian             & 0.83{\small $\pm$0.72}                  & 15.00{\small $\pm$1.25}                    & 20.83{\small $\pm$0.72}                   & 62.50{\small $\pm$0.00}   & 80.42{\small $\pm$3.61} \\
Indo-European (Germanic) & 1.25{\small $\pm$1.77}                  & 15.94{\small $\pm$2.58}                    & 20.94{\small $\pm$2.13}                   & 64.06{\small $\pm$2.77}   & 80.31{\small $\pm$1.57} \\
Indo-European (Romance)  & 0.31{\small $\pm$0.62}                  & 15.31{\small $\pm$1.57}                    & 20.31{\small $\pm$0.63}                   & 66.25{\small $\pm$3.68}   & 79.06{\small $\pm$1.57} \\
Indo-European (Greek)    & 1.25{\small $\pm$0.00}                  & 12.50{\small $\pm$0.00}                    & 17.50{\small $\pm$0.00}                   & 53.75{\small $\pm$0.00}   & 71.25{\small $\pm$0.00} \\
Indo-European (Iranian)  & 0.00{\small $\pm$0.00}                  & 12.50{\small $\pm$0.00}                    & 17.50{\small $\pm$0.00}                   & 60.00{\small $\pm$0.00}   & 78.75{\small $\pm$0.00} \\
Slavic                   & 0.63{\small $\pm$0.88}                  & 14.38{\small $\pm$0.88}                    & 18.13{\small $\pm$0.88}                   & 66.88{\small $\pm$7.95}   & 74.38{\small $\pm$0.88} \\
Sino-Tibetan             & 0.00{\small $\pm$0.00}                  & 15.00{\small $\pm$0.00}                    & 20.00{\small $\pm$0.00}                   & 65.00{\small $\pm$0.00}   & 78.75{\small $\pm$0.00} \\
Turkic                   & 1.25{\small $\pm$0.00}                  & 15.00{\small $\pm$0.00}                    & 18.75{\small $\pm$0.00}                   & 62.50{\small $\pm$0.00}   & 73.75{\small $\pm$0.00} \\
Uralic                   & 1.25{\small $\pm$1.25}                  & 14.17{\small $\pm$3.61}                    & 19.58{\small $\pm$4.39}                   & 62.50{\small $\pm$4.51}   & 79.58{\small $\pm$5.20} \\ \bottomrule
\end{tabular}%
}
\caption{Performance comparison of different models on Python across language families. Pass@1 results are reported, with scores averaged across different languages within each group as well as standard deviation.}
\label{tab:lang_family_result}
\end{table}

The difference in pass@1 among categorized groups is not large yet consistent, with class 3 consistently demonstrating lower performance across different models. We further conduct similar experiments on different NL families. Table \ref{tab:lang_family_result} highlights that languages from the Afro-Asiatic, Indo-European (Greek), Indo-European (Iranian), and Turkic language families generally yield lower results compared to other language families. These findings underscore a significant challenge: Given NL prompts expressing the same meaning in different languages, current LLMs struggle to capture the equivalent semantic meaning.


\section{Conclusion}
We propose {\ours}, a massively multilingual code generation benchmark for assessing cross-lingual NL generalization for LLMs. Drawing inspiration from back-translation, we have devised an iterative process that enables the creation of a robust benchmark spanning 12 PLs and 23 NLs. Our study underscores the persistent challenge in achieving effective cross-lingual NL generalization, highlighting a key area for future research.

\section*{Acknowledgments}
We thank all anonymous reviewers for their insightful comments and feedback. Their expertise and thoughtful critique have significantly contributed to the refinement of our work. This work was partially supported by DisAI - Improving scientific excellence and creativity in combating disinformation with artificial intelligence and language technologies, a project funded by European Union under the Horizon Europe, GA No. 101079164.

\section*{Bibliographical References}\label{sec:reference}

\bibliographystyle{lrec-coling2024-natbib}
\bibliography{lrec-coling2024-example}

\bibliographystylelanguageresource{lrec-coling2024-natbib}
\bibliographylanguageresource{languageresource}

\appendix
\clearpage

\section{Experiment Settings}

\paragraph{Inference Hyperparameter}
In our experimental framework, we consistently employed top-$p$ sampling across all evaluated models. The specific parameters for the sampling process were set to a top-$p$ value of 0.95 and a temperature of 0.2. This uniform approach ensures comparability of results across different models.

\paragraph{Evaluated Models}
The models included in our analysis span a range of complexities to adequately assess the impact of model size on performance. Specifically, we evaluated:
\begin{itemize}
    \item \textbf{CodeGen2} models with sizes of 1B, 3.7B, 7B, and 16B parameters.
    \item \textbf{CodeT5+} variants with 220M, 770M, and 2B parameters.
    \item OpenAI \textbf{GPT-3.5} and \textbf{GPT-4} models.
\end{itemize}

This diverse set of models allows for a comprehensive examination of the relationship between model size and program synthesis task within our evaluation benchmark. Following \citet{mxeval23}, we keep only the first generated function as our stopping criteria.


\section{Comprehensive Experiment Results}
In the appendix, we report our full experimental results across different models with varying sizes, including CodeGen2 (1B, 3.7B, 7B, 16B), CodeT5+ (220M, 770M, 2B) and GPT (3.5, 4) in following tables

Detailed results for each model, across different programming languages, are meticulously documented in the subsequent tables. These results provide a holistic view of our analysis, allowing for an in-depth comparison of model performances.

\begin{itemize}
    \item Python programming language results are provided in 
Table~\ref{tab:appendix_python}.
    \item Java results are detailed in Table~\ref{tab:appendix_java}.
    \item JavaScript findings are recorded in Table~\ref{tab:appendix_javascript}.
    \item C\# outcomes are laid out in Table~\ref{tab:appendix_csharp}.
   \item Go performance metrics are available in Table~\ref{tab:appendix_go}.
    \item Kotlin results can be found in Table~\ref{tab:appendix_kotlin}.
    \item Perl evaluation is presented in Table~\ref{tab:appendix_perl}.
    \item PHP findings are described in Table~\ref{tab:appendix_php}.
    \item Ruby results are listed in Table~\ref{tab:appendix_ruby}.
    \item Scala outcomes are outlined in Table~\ref{tab:appendix_scala}.
    \item Swift performance analysis is provided in Table~\ref{tab:appendix_swift}.
    \item TypeScript results are compiled in Table~\ref{tab:appendix_typescript}.
\end{itemize}

Through these tables, readers can access the full scope of our experimental findings, enabling a comprehensive understanding of how each model performs across various programming languages.

\begin{table*}[]
    \centering
    \resizebox{0.9\textwidth}{!}{%
    \begin{tabular}{llllllllll}
    \toprule
    {} & CodeT5+(220M) & CodeT5+(770M) & CodeT5+(2B) & CodeGen2(1B) & CodeGen2(3.7B) & CodeGen2(7B) & CodeGen2(16B) & GPT-3.5 &  GPT-4 \\
    \midrule
    English    &          0.00 &          2.50 &        3.75 &        11.25 &          18.75 &        21.25 &         23.75 &   66.25 &  82.50 \\
    Russian    &          0.00 &          0.00 &        1.25 &         8.75 &          15.00 &        18.75 &         17.50 &   72.50 &  73.75 \\
    Chinese    &          0.00 &          0.00 &        1.25 &         8.75 &          15.00 &        21.25 &         20.00 &   65.00 &  78.75 \\
    German     &          0.00 &          0.00 &        1.25 &         8.75 &          16.25 &        18.75 &         21.25 &   65.00 &  80.00 \\
    Spanish    &          0.00 &          1.25 &        1.25 &         8.75 &          15.00 &        18.75 &         20.00 &   63.75 &  78.75 \\
    French     &          0.00 &          1.25 &        1.25 &         8.75 &          13.75 &        21.25 &         20.00 &   62.50 &  77.50 \\
    Italian    &          0.00 &          0.00 &        1.25 &        12.50 &          15.00 &        17.50 &         20.00 &   70.00 &  81.25 \\
    Portuguese &          0.00 &          0.00 &        1.25 &        11.25 &          17.50 &        21.25 &         21.25 &   68.75 &  78.75 \\
    Greek      &          0.00 &          0.00 &        1.25 &        11.25 &          12.50 &        18.75 &         17.50 &   53.75 &  71.25 \\
    Hungarian  &          0.00 &          0.00 &        1.25 &        10.00 &          10.00 &        15.00 &         15.00 &   67.50 &  83.75 \\
    Dutch      &          0.00 &          0.00 &        1.25 &        10.00 &          16.25 &        22.50 &         18.75 &   65.00 &  78.75 \\
    Finnish    &          0.00 &          0.00 &        2.50 &        10.00 &          16.25 &        15.00 &         23.75 &   61.25 &  73.75 \\
    Indonesian &          0.00 &          0.00 &        1.25 &        10.00 &          16.25 &        17.50 &         21.25 &   62.50 &  76.25 \\
    Turkish    &          0.00 &          0.00 &        1.25 &         7.50 &          15.00 &        20.00 &         18.75 &   62.50 &  73.75 \\
    Arabic     &          0.00 &          0.00 &        1.25 &         8.75 &          13.75 &        16.25 &         20.00 &   52.50 &  73.75 \\
    Vietnamese &          0.00 &          0.00 &        1.25 &         7.50 &          15.00 &        20.00 &         18.75 &   66.25 &  81.25 \\
    Bulgarian  &          0.00 &          0.00 &        1.25 &         8.75 &          13.75 &        18.75 &         18.75 &   61.25 &  75.00 \\
    Persian    &          0.00 &          0.00 &        1.25 &        11.25 &          12.50 &        16.25 &         17.50 &   60.00 &  78.75 \\
    Malay      &          0.00 &          0.00 &        1.25 &         8.75 &          13.75 &        18.75 &         21.25 &   62.50 &  82.50 \\
    Hebrew     &          0.00 &          1.25 &        1.25 &        10.00 &          13.75 &        18.75 &         18.75 &   60.00 &  76.25 \\
    Estonian   &          0.00 &          0.00 &        1.25 &        11.25 &          16.25 &        16.25 &         20.00 &   58.75 &  81.25 \\
    Tagalog    &          0.00 &          0.00 &        1.25 &         7.50 &          15.00 &        16.25 &         20.00 &   62.50 &  82.50 \\
    Afrikaans  &          0.00 &          0.00 &        1.25 &         8.75 &          12.50 &        17.50 &         20.00 &   60.00 &  80.00 \\
    \bottomrule
    \end{tabular}
    }

    \caption{Results (pass@1) of different models on Python across 23 natural languages.}
    \label{tab:appendix_python}
\end{table*}

\begin{table*}[]
    \centering
    \resizebox{0.9\textwidth}{!}{%
    \begin{tabular}{llllllllll}
    \toprule
    {} & CodeT5+(220M) & CodeT5+(770M) & CodeT5+(2B) & CodeGen2(1B) & CodeGen2(3.7B) & CodeGen2(7B) & CodeGen2(16B) & GPT-3.5 &  GPT-4 \\
    \midrule
    English    &          0.00 &          1.25 &        0.00 &        12.50 &          16.25 &        18.75 &         25.00 &    1.25 &  67.50 \\
    Russian    &          0.00 &          1.25 &        0.00 &        10.00 &           8.75 &        17.50 &         20.00 &    0.00 &  66.25 \\
    Chinese    &          0.00 &          0.00 &        0.00 &        10.00 &          13.75 &        20.00 &         18.75 &    0.00 &  63.75 \\
    German     &          0.00 &          1.25 &        0.00 &         8.75 &          12.50 &        16.25 &         20.00 &    0.00 &  62.50 \\
    Spanish    &          0.00 &          1.25 &        0.00 &        11.25 &          11.25 &        17.50 &         23.75 &    0.00 &  62.50 \\
    French     &          0.00 &          1.25 &        0.00 &         7.50 &          13.75 &        18.75 &         20.00 &    0.00 &  57.50 \\
    Italian    &          0.00 &          1.25 &        0.00 &        11.25 &          12.50 &        17.50 &         18.75 &    0.00 &  63.75 \\
    Portuguese &          0.00 &          0.00 &        0.00 &         7.50 &          10.00 &        18.75 &         25.00 &    0.00 &  72.50 \\
    Greek      &          0.00 &          2.50 &        0.00 &         7.50 &          11.25 &        17.50 &         15.00 &    0.00 &  58.75 \\
    Hungarian  &          0.00 &          1.25 &        0.00 &         7.50 &          16.25 &        16.25 &         17.50 &    0.00 &  56.25 \\
    Dutch      &          0.00 &          1.25 &        0.00 &        10.00 &          15.00 &        16.25 &         17.50 &    2.50 &  63.75 \\
    Finnish    &          0.00 &          0.00 &        0.00 &         7.50 &          15.00 &        13.75 &         18.75 &    0.00 &  62.50 \\
    Indonesian &          0.00 &          1.25 &        0.00 &         8.75 &          12.50 &        15.00 &         21.25 &    1.25 &  58.75 \\
    Turkish    &          0.00 &          0.00 &        0.00 &         8.75 &          10.00 &        11.25 &         18.75 &    0.00 &  61.25 \\
    Arabic     &          0.00 &          0.00 &        0.00 &         7.50 &           8.75 &        13.75 &         16.25 &    0.00 &  53.75 \\
    Vietnamese &          0.00 &          1.25 &        0.00 &         7.50 &          11.25 &        16.25 &         20.00 &    1.25 &  60.00 \\
    Bulgarian  &          0.00 &          1.25 &        0.00 &        10.00 &          10.00 &        20.00 &         18.75 &    0.00 &  60.00 \\
    Persian    &          0.00 &          1.25 &        0.00 &         7.50 &           8.75 &        16.25 &         15.00 &    0.00 &  61.25 \\
    Malay      &          0.00 &          1.25 &        0.00 &         7.50 &          11.25 &        15.00 &         18.75 &    1.25 &  61.25 \\
    Hebrew     &          0.00 &          0.00 &        0.00 &         6.25 &          12.50 &        13.75 &         16.25 &    0.00 &  66.25 \\
    Estonian   &          0.00 &          1.25 &        0.00 &        10.00 &          11.25 &        13.75 &         15.00 &    0.00 &  55.00 \\
    Tagalog    &          0.00 &          1.25 &        0.00 &         8.75 &           8.75 &        12.50 &         17.50 &    0.00 &  70.00 \\
    Afrikaans  &          0.00 &          1.25 &        0.00 &         7.50 &          12.50 &        15.00 &         16.25 &    0.00 &  58.75 \\
    \bottomrule
    \end{tabular}

    }

    \caption{Results (pass@1) of different models on Java across 23 natural languages.}
    \label{tab:appendix_java}
\end{table*}

\begin{table*}[]
    \centering
    \resizebox{0.9\textwidth}{!}{%
    \begin{tabular}{llllllllll}
    \toprule
    {} & CodeT5+(220M) & CodeT5+(770M) & CodeT5+(2B) & CodeGen2(1B) & CodeGen2(3.7B) & CodeGen2(7B) & CodeGen2(16B) & GPT-3.5 &  GPT-4 \\
    \midrule
    English    &          0.00 &          0.00 &        0.00 &        11.25 &          12.50 &        17.50 &         20.00 &    3.75 &  80.00 \\
    Russian    &          0.00 &          2.50 &        0.00 &         6.25 &          17.50 &        16.25 &         21.25 &    2.50 &  81.25 \\
    Chinese    &          0.00 &          2.50 &        0.00 &        10.00 &          17.50 &        17.50 &         23.75 &    3.75 &  76.25 \\
    German     &          0.00 &          0.00 &        0.00 &         6.25 &          12.50 &        20.00 &         20.00 &    7.50 &  76.25 \\
    Spanish    &          0.00 &          2.50 &        0.00 &        13.75 &          12.50 &        16.25 &         23.75 &    8.75 &  75.00 \\
    French     &          0.00 &          1.25 &        0.00 &        10.00 &          12.50 &        21.25 &         21.25 &    6.25 &  72.50 \\
    Italian    &          0.00 &          1.25 &        0.00 &         8.75 &          12.50 &        16.25 &         23.75 &    8.75 &  78.75 \\
    Portuguese &          0.00 &          2.50 &        0.00 &        12.50 &          12.50 &        17.50 &         21.25 &    8.75 &  81.25 \\
    Greek      &          0.00 &          2.50 &        0.00 &         7.50 &          15.00 &        17.50 &         20.00 &    1.25 &  70.00 \\
    Hungarian  &          0.00 &          0.00 &        0.00 &        10.00 &          11.25 &        16.25 &         21.25 &    7.50 &  81.25 \\
    Dutch      &          0.00 &          0.00 &        0.00 &        11.25 &          17.50 &        21.25 &         21.25 &    8.75 &  75.00 \\
    Finnish    &          0.00 &          1.25 &        0.00 &         7.50 &          12.50 &        15.00 &         20.00 &    5.00 &  77.50 \\
    Indonesian &          0.00 &          0.00 &        0.00 &         8.75 &          15.00 &        17.50 &         21.25 &    5.00 &  75.00 \\
    Turkish    &          0.00 &          0.00 &        0.00 &         8.75 &          13.75 &        18.75 &         21.25 &    2.50 &  73.75 \\
    Arabic     &          0.00 &          2.50 &        0.00 &         7.50 &          15.00 &        12.50 &         16.25 &    1.25 &  68.75 \\
    Vietnamese &          0.00 &          3.75 &        0.00 &        12.50 &          18.75 &        15.00 &         22.50 &    5.00 &  76.25 \\
    Bulgarian  &          0.00 &          3.75 &        0.00 &        10.00 &          11.25 &        17.50 &         23.75 &    1.25 &  78.75 \\
    Persian    &          0.00 &          3.75 &        0.00 &         7.50 &          16.25 &        16.25 &         21.25 &    3.75 &  73.75 \\
    Malay      &          0.00 &          0.00 &        0.00 &         8.75 &          12.50 &        16.25 &         25.00 &    1.25 &  73.75 \\
    Hebrew     &          0.00 &          1.25 &        0.00 &         8.75 &          15.00 &        16.25 &         18.75 &    2.50 &  72.50 \\
    Estonian   &          0.00 &          0.00 &        0.00 &        11.25 &          11.25 &        16.25 &         22.50 &    2.50 &  78.75 \\
    Tagalog    &          0.00 &          0.00 &        0.00 &         8.75 &          15.00 &        16.25 &         18.75 &    6.25 &  77.50 \\
    Afrikaans  &          0.00 &          2.50 &        0.00 &         8.75 &          11.25 &        17.50 &         21.25 &    6.25 &  77.50 \\
    \bottomrule
    \end{tabular}

    }

    \caption{Results (pass@1) of different models on JavaScript across 23 natural languages.}
    \label{tab:appendix_javascript}
\end{table*}

\begin{table*}[]
    \centering
    \resizebox{0.9\textwidth}{!}{%
    \begin{tabular}{llllllllll}
    \toprule
    {} & CodeT5+(220M) & CodeT5+(770M) & CodeT5+(2B) & CodeGen2(1B) & CodeGen2(3.7B) & CodeGen2(7B) & CodeGen2(16B) & GPT-3.5 &  GPT-4 \\
    \midrule
    English    &          0.00 &          3.75 &        0.00 &         8.75 &          10.00 &        18.75 &         22.50 &    2.50 &  51.25 \\
    Russian    &          0.00 &          1.25 &        0.00 &         6.25 &           8.75 &        17.50 &         18.75 &    3.75 &  60.00 \\
    Chinese    &          0.00 &          1.25 &        0.00 &         7.50 &          12.50 &        13.75 &         21.25 &    3.75 &  57.50 \\
    German     &          0.00 &          2.50 &        0.00 &         8.75 &          10.00 &        13.75 &         20.00 &    5.00 &  62.50 \\
    Spanish    &          0.00 &          3.75 &        0.00 &        10.00 &          10.00 &        17.50 &         22.50 &    3.75 &  63.75 \\
    French     &          0.00 &          2.50 &        0.00 &        10.00 &          11.25 &        16.25 &         21.25 &    3.75 &  57.50 \\
    Italian    &          0.00 &          2.50 &        0.00 &         8.75 &           8.75 &        15.00 &         25.00 &    3.75 &  56.25 \\
    Portuguese &          0.00 &          2.50 &        0.00 &        10.00 &          11.25 &        15.00 &         23.75 &    6.25 &  63.75 \\
    Greek      &          0.00 &          1.25 &        0.00 &         6.25 &          10.00 &        12.50 &         13.75 &    2.50 &  53.75 \\
    Hungarian  &          0.00 &          3.75 &        0.00 &         7.50 &          10.00 &        12.50 &         12.50 &    6.25 &  57.50 \\
    Dutch      &          0.00 &          2.50 &        0.00 &         7.50 &          10.00 &        16.25 &         22.50 &    7.50 &  56.25 \\
    Finnish    &          0.00 &          1.25 &        0.00 &         6.25 &          10.00 &        13.75 &         15.00 &    2.50 &  56.25 \\
    Indonesian &          0.00 &          2.50 &        0.00 &         7.50 &          10.00 &        11.25 &         21.25 &    2.50 &  56.25 \\
    Turkish    &          0.00 &          1.25 &        0.00 &         7.50 &          11.25 &        13.75 &         23.75 &    3.75 &  55.00 \\
    Arabic     &          0.00 &          0.00 &        0.00 &         6.25 &          11.25 &        13.75 &         21.25 &    5.00 &  56.25 \\
    Vietnamese &          0.00 &          3.75 &        0.00 &         7.50 &           8.75 &        13.75 &         21.25 &    3.75 &  58.75 \\
    Bulgarian  &          0.00 &          5.00 &        0.00 &         8.75 &          11.25 &        15.00 &         18.75 &    3.75 &  58.75 \\
    Persian    &          0.00 &          1.25 &        0.00 &         6.25 &          11.25 &        13.75 &         16.25 &    3.75 &  57.50 \\
    Malay      &          0.00 &          2.50 &        0.00 &         7.50 &          11.25 &        12.50 &         20.00 &    1.25 &  60.00 \\
    Hebrew     &          0.00 &          1.25 &        0.00 &         5.00 &          11.25 &        13.75 &         16.25 &    3.75 &  53.75 \\
    Estonian   &          0.00 &          2.50 &        0.00 &         6.25 &          11.25 &        13.75 &         17.50 &    3.75 &  56.25 \\
    Tagalog    &          0.00 &          1.25 &        0.00 &         7.50 &          10.00 &        12.50 &         17.50 &    2.50 &  63.75 \\
    Afrikaans  &          0.00 &          2.50 &        0.00 &         7.50 &           8.75 &        13.75 &         17.50 &    6.25 &  52.50 \\
    \bottomrule
    \end{tabular}

    }

    \caption{Results (pass@1) of different models on C\# across 23 natural languages.}
    \label{tab:appendix_csharp}
\end{table*}

\begin{table*}[]
    \centering
    \resizebox{0.9\textwidth}{!}{%
    \begin{tabular}{llllllllll}
    \toprule
    {} & CodeT5+(220M) & CodeT5+(770M) & CodeT5+(2B) & CodeGen2(1B) & CodeGen2(3.7B) & CodeGen2(7B) & CodeGen2(16B) & GPT-3.5 &  GPT-4 \\
    \midrule
    English    &          0.00 &          1.25 &        0.00 &         2.50 &          10.00 &        12.50 &         13.75 &    2.50 &  47.50 \\
    Russian    &          0.00 &          0.00 &        0.00 &         3.75 &          10.00 &        15.00 &         15.00 &    5.00 &  50.00 \\
    Chinese    &          0.00 &          1.25 &        0.00 &         3.75 &          10.00 &        17.50 &         18.75 &    2.50 &  47.50 \\
    German     &          0.00 &          1.25 &        0.00 &         2.50 &           7.50 &        11.25 &         16.25 &    2.50 &  41.25 \\
    Spanish    &          0.00 &          0.00 &        0.00 &         3.75 &           7.50 &        12.50 &         13.75 &    3.75 &  45.00 \\
    French     &          0.00 &          0.00 &        0.00 &         2.50 &           7.50 &        16.25 &         12.50 &    2.50 &  43.75 \\
    Italian    &          0.00 &          0.00 &        0.00 &         3.75 &           7.50 &        16.25 &         15.00 &    3.75 &  48.75 \\
    Portuguese &          0.00 &          1.25 &        0.00 &         2.50 &          11.25 &        15.00 &         16.25 &    3.75 &  45.00 \\
    Greek      &          0.00 &          0.00 &        0.00 &         2.50 &           7.50 &        11.25 &         15.00 &    6.25 &  41.25 \\
    Hungarian  &          0.00 &          0.00 &        0.00 &         3.75 &           8.75 &        13.75 &         13.75 &    3.75 &  45.00 \\
    Dutch      &          0.00 &          1.25 &        0.00 &         3.75 &           7.50 &        13.75 &         15.00 &    2.50 &  50.00 \\
    Finnish    &          0.00 &          1.25 &        0.00 &         2.50 &           7.50 &        12.50 &         16.25 &    1.25 &  43.75 \\
    Indonesian &          0.00 &          1.25 &        0.00 &         2.50 &           7.50 &        13.75 &         17.50 &    1.25 &  38.75 \\
    Turkish    &          0.00 &          1.25 &        0.00 &         3.75 &           6.25 &        11.25 &         16.25 &    5.00 &  45.00 \\
    Arabic     &          0.00 &          1.25 &        0.00 &         3.75 &           6.25 &         8.75 &         15.00 &    3.75 &  43.75 \\
    Vietnamese &          0.00 &          1.25 &        0.00 &         3.75 &           7.50 &        13.75 &         17.50 &    2.50 &  43.75 \\
    Bulgarian  &          0.00 &          0.00 &        0.00 &         3.75 &           7.50 &        15.00 &         15.00 &    7.50 &  47.50 \\
    Persian    &          0.00 &          1.25 &        0.00 &         2.50 &           7.50 &        12.50 &         13.75 &    5.00 &  45.00 \\
    Malay      &          0.00 &          1.25 &        0.00 &         3.75 &           7.50 &        11.25 &         17.50 &    5.00 &  47.50 \\
    Hebrew     &          0.00 &          0.00 &        0.00 &         2.50 &           7.50 &        12.50 &         16.25 &    6.25 &  45.00 \\
    Estonian   &          0.00 &          1.25 &        0.00 &         5.00 &           6.25 &        13.75 &         17.50 &    6.25 &  43.75 \\
    Tagalog    &          0.00 &          1.25 &        0.00 &         2.50 &           6.25 &        10.00 &         15.00 &    2.50 &  40.00 \\
    Afrikaans  &          0.00 &          0.00 &        0.00 &         3.75 &           7.50 &        11.25 &         18.75 &    3.75 &  41.25 \\
    \bottomrule
    \end{tabular}

    }

    \caption{Results (pass@1) of different models on Go across 23 natural languages.}
    \label{tab:appendix_go}
\end{table*}

\begin{table*}[]
    \centering
    \resizebox{0.9\textwidth}{!}{%
    \begin{tabular}{llllllllll}
    \toprule
    {} & CodeT5+(220M) & CodeT5+(770M) & CodeT5+(2B) & CodeGen2(1B) & CodeGen2(3.7B) & CodeGen2(7B) & CodeGen2(16B) & GPT-3.5 &  GPT-4 \\
    \midrule
    English    &          0.00 &          0.00 &        0.00 &         5.00 &           8.75 &        18.75 &         18.75 &    7.50 &  63.75 \\
    Russian    &          0.00 &          1.25 &        0.00 &         3.75 &           8.75 &        13.75 &         20.00 &    7.50 &  68.75 \\
    Chinese    &          0.00 &          0.00 &        0.00 &         6.25 &           3.75 &        20.00 &         16.25 &    6.25 &  67.50 \\
    German     &          0.00 &          0.00 &        0.00 &         5.00 &          10.00 &        17.50 &         17.50 &    8.75 &  68.75 \\
    Spanish    &          0.00 &          0.00 &        0.00 &         1.25 &           7.50 &        16.25 &         20.00 &   11.25 &  65.00 \\
    French     &          0.00 &          0.00 &        0.00 &         3.75 &           8.75 &        12.50 &         18.75 &    7.50 &  67.50 \\
    Italian    &          0.00 &          0.00 &        0.00 &         3.75 &           7.50 &        16.25 &         16.25 &    7.50 &  68.75 \\
    Portuguese &          0.00 &          0.00 &        0.00 &         2.50 &           3.75 &        16.25 &         18.75 &    5.00 &  70.00 \\
    Greek      &          0.00 &          1.25 &        0.00 &         5.00 &           7.50 &        12.50 &         12.50 &    3.75 &  63.75 \\
    Hungarian  &          0.00 &          0.00 &        0.00 &         2.50 &           6.25 &        18.75 &         12.50 &    6.25 &  73.75 \\
    Dutch      &          0.00 &          0.00 &        0.00 &         5.00 &           7.50 &        15.00 &         16.25 &    6.25 &  63.75 \\
    Finnish    &          0.00 &          0.00 &        0.00 &         2.50 &           8.75 &        16.25 &         12.50 &    8.75 &  65.00 \\
    Indonesian &          0.00 &          0.00 &        0.00 &         3.75 &           7.50 &        16.25 &         17.50 &    8.75 &  62.50 \\
    Turkish    &          0.00 &          0.00 &        0.00 &         5.00 &           6.25 &        11.25 &         15.00 &    7.50 &  61.25 \\
    Arabic     &          0.00 &          1.25 &        0.00 &         3.75 &           7.50 &        16.25 &         13.75 &    7.50 &  61.25 \\
    Vietnamese &          0.00 &          1.25 &        0.00 &         3.75 &           7.50 &        15.00 &         18.75 &    5.00 &  65.00 \\
    Bulgarian  &          0.00 &          1.25 &        0.00 &         5.00 &           7.50 &        15.00 &         16.25 &    6.25 &  67.50 \\
    Persian    &          0.00 &          1.25 &        0.00 &         3.75 &           6.25 &        13.75 &         16.25 &    3.75 &  65.00 \\
    Malay      &          0.00 &          0.00 &        0.00 &         2.50 &           5.00 &        13.75 &         15.00 &    8.75 &  63.75 \\
    Hebrew     &          0.00 &          1.25 &        0.00 &         5.00 &           5.00 &        15.00 &         12.50 &    2.50 &  61.25 \\
    Estonian   &          0.00 &          0.00 &        0.00 &         3.75 &           8.75 &        15.00 &         15.00 &    8.75 &  65.00 \\
    Tagalog    &          0.00 &          0.00 &        0.00 &         5.00 &           8.75 &        15.00 &         12.50 &    8.75 &  66.25 \\
    Afrikaans  &          0.00 &          0.00 &        0.00 &         3.75 &           5.00 &        13.75 &         17.50 &    8.75 &  66.25 \\
    \bottomrule
    \end{tabular}

    }

    \caption{Results (pass@1) of different models on Kotlin across 23 natural languages.}
    \label{tab:appendix_kotlin}
\end{table*}

\begin{table*}[]
    \centering
    \resizebox{0.9\textwidth}{!}{%
    \begin{tabular}{llllllllll}
    \toprule
    {} & CodeT5+(220M) & CodeT5+(770M) & CodeT5+(2B) & CodeGen2(1B) & CodeGen2(3.7B) & CodeGen2(7B) & CodeGen2(16B) & GPT-3.5 &  GPT-4 \\
    \midrule
    English    &          0.00 &          0.00 &        0.00 &         0.00 &           2.50 &         1.25 &          2.50 &   16.25 &  62.50 \\
    Russian    &          0.00 &          0.00 &        0.00 &         1.25 &           0.00 &         1.25 &          5.00 &   16.25 &  46.25 \\
    Chinese    &          0.00 &          0.00 &        0.00 &         2.50 &           1.25 &         1.25 &          3.75 &   18.75 &  52.50 \\
    German     &          0.00 &          0.00 &        0.00 &         1.25 &           2.50 &         1.25 &          1.25 &   25.00 &  55.00 \\
    Spanish    &          0.00 &          0.00 &        0.00 &         1.25 &           1.25 &         3.75 &          2.50 &   25.00 &  52.50 \\
    French     &          0.00 &          0.00 &        0.00 &         0.00 &           2.50 &         2.50 &          1.25 &   22.50 &  48.75 \\
    Italian    &          0.00 &          0.00 &        0.00 &         0.00 &           0.00 &         2.50 &          2.50 &   26.25 &  52.50 \\
    Portuguese &          0.00 &          0.00 &        0.00 &         0.00 &           2.50 &         0.00 &          3.75 &   21.25 &  56.25 \\
    Greek      &          0.00 &          0.00 &        0.00 &         1.25 &           1.25 &         2.50 &          0.00 &   10.00 &  48.75 \\
    Hungarian  &          0.00 &          0.00 &        0.00 &         1.25 &           1.25 &         1.25 &          0.00 &   27.50 &  57.50 \\
    Dutch      &          0.00 &          0.00 &        0.00 &         2.50 &           1.25 &         1.25 &          3.75 &   26.25 &  50.00 \\
    Finnish    &          0.00 &          0.00 &        0.00 &         0.00 &           1.25 &         1.25 &          1.25 &   18.75 &  52.50 \\
    Indonesian &          0.00 &          0.00 &        0.00 &         2.50 &           1.25 &         0.00 &          1.25 &   17.50 &  55.00 \\
    Turkish    &          0.00 &          0.00 &        0.00 &         0.00 &           0.00 &         1.25 &          2.50 &   23.75 &  51.25 \\
    Arabic     &          0.00 &          0.00 &        0.00 &         1.25 &           1.25 &         0.00 &          1.25 &   18.75 &  50.00 \\
    Vietnamese &          0.00 &          0.00 &        0.00 &         1.25 &           2.50 &         1.25 &          1.25 &   12.50 &  52.50 \\
    Bulgarian  &          0.00 &          0.00 &        0.00 &         1.25 &           1.25 &         2.50 &          1.25 &   20.00 &  50.00 \\
    Persian    &          0.00 &          0.00 &        0.00 &         1.25 &           0.00 &         1.25 &          1.25 &   23.75 &  52.50 \\
    Malay      &          0.00 &          0.00 &        0.00 &         1.25 &           2.50 &         1.25 &          1.25 &   16.25 &  55.00 \\
    Hebrew     &          0.00 &          0.00 &        0.00 &         0.00 &           0.00 &         2.50 &          3.75 &   11.25 &  48.75 \\
    Estonian   &          0.00 &          0.00 &        0.00 &         0.00 &           0.00 &         1.25 &          2.50 &   21.25 &  52.50 \\
    Tagalog    &          0.00 &          0.00 &        0.00 &         0.00 &           0.00 &         2.50 &          2.50 &   18.75 &  52.50 \\
    Afrikaans  &          0.00 &          0.00 &        0.00 &         1.25 &           3.75 &         1.25 &          1.25 &   20.00 &  50.00 \\
    \bottomrule
    \end{tabular}

    }

    \caption{Results (pass@1) of different models on Perl across 23 natural languages.}
    \label{tab:appendix_perl}
\end{table*}

\begin{table*}[]
    \centering
    \resizebox{0.9\textwidth}{!}{%
    \begin{tabular}{llllllllll}
    \toprule
    {} & CodeT5+(220M) & CodeT5+(770M) & CodeT5+(2B) & CodeGen2(1B) & CodeGen2(3.7B) & CodeGen2(7B) & CodeGen2(16B) & GPT-3.5 &  GPT-4 \\
    \midrule
    English    &          0.00 &         26.25 &        0.00 &         7.50 &          13.75 &        20.00 &         22.50 &    2.50 &  78.75 \\
    Russian    &          1.25 &          7.50 &        0.00 &         3.75 &          15.00 &        16.25 &         21.25 &    6.25 &  73.75 \\
    Chinese    &          0.00 &         13.75 &        0.00 &         6.25 &          17.50 &        20.00 &         21.25 &    2.50 &  72.50 \\
    German     &          0.00 &         17.50 &        0.00 &         3.75 &          10.00 &        17.50 &         17.50 &    8.75 &  78.75 \\
    Spanish    &          0.00 &         21.25 &        0.00 &         7.50 &          16.25 &        20.00 &         22.50 &    3.75 &  76.25 \\
    French     &          0.00 &         17.50 &        0.00 &         6.25 &          15.00 &        16.25 &         20.00 &    2.50 &  76.25 \\
    Italian    &          0.00 &         16.25 &        0.00 &         5.00 &          17.50 &        15.00 &         21.25 &    3.75 &  82.50 \\
    Portuguese &          0.00 &         13.75 &        0.00 &         5.00 &          13.75 &        15.00 &         21.25 &    2.50 &  78.75 \\
    Greek      &          0.00 &          8.75 &        0.00 &         3.75 &          15.00 &        12.50 &         17.50 &    3.75 &  75.00 \\
    Hungarian  &          0.00 &          1.25 &        0.00 &         3.75 &          11.25 &        13.75 &         18.75 &    6.25 &  80.00 \\
    Dutch      &          0.00 &         17.50 &        0.00 &         2.50 &          11.25 &        20.00 &         18.75 &    6.25 &  75.00 \\
    Finnish    &          0.00 &          1.25 &        0.00 &         3.75 &          11.25 &        13.75 &         20.00 &    6.25 &  76.25 \\
    Indonesian &          0.00 &          5.00 &        0.00 &         2.50 &          12.50 &        15.00 &         21.25 &    2.50 &  78.75 \\
    Turkish    &          0.00 &          3.75 &        0.00 &         3.75 &          12.50 &        13.75 &         18.75 &    5.00 &  73.75 \\
    Arabic     &          1.25 &          7.50 &        0.00 &         3.75 &          12.50 &        13.75 &         17.50 &    5.00 &  75.00 \\
    Vietnamese &          0.00 &          5.00 &        0.00 &         2.50 &          11.25 &        16.25 &         17.50 &    0.00 &  77.50 \\
    Bulgarian  &          2.50 &          7.50 &        0.00 &         2.50 &          12.50 &        18.75 &         17.50 &    1.25 &  81.25 \\
    Persian    &          0.00 &         10.00 &        0.00 &         3.75 &          11.25 &        13.75 &         16.25 &    1.25 &  73.75 \\
    Malay      &          0.00 &          6.25 &        0.00 &         5.00 &          12.50 &        15.00 &         20.00 &    2.50 &  76.25 \\
    Hebrew     &          0.00 &         15.00 &        0.00 &         3.75 &          13.75 &        12.50 &         18.75 &    3.75 &  73.75 \\
    Estonian   &          0.00 &          8.75 &        0.00 &         3.75 &          11.25 &        15.00 &         18.75 &    3.75 &  78.75 \\
    Tagalog    &          0.00 &          2.50 &        0.00 &         3.75 &          13.75 &        12.50 &         17.50 &    2.50 &  76.25 \\
    Afrikaans  &          0.00 &         13.75 &        0.00 &         3.75 &          13.75 &        11.25 &         16.25 &    6.25 &  78.75 \\
    \bottomrule
    \end{tabular}

    }

    \caption{Results (pass@1) of different models on PHP across 23 natural languages.}
    \label{tab:appendix_php}
\end{table*}

\begin{table*}[]
    \centering
    \resizebox{0.9\textwidth}{!}{%
    \begin{tabular}{llllllllll}
    \toprule
    {} & CodeT5+(220M) & CodeT5+(770M) & CodeT5+(2B) & CodeGen2(1B) & CodeGen2(3.7B) & CodeGen2(7B) & CodeGen2(16B) & GPT-3.5 &  GPT-4 \\
    \midrule
    English    &          0.00 &          0.00 &        0.00 &         3.75 &          11.25 &        17.50 &         13.75 &   11.25 &  66.25 \\
    Russian    &          0.00 &          0.00 &        0.00 &         7.50 &           8.75 &        11.25 &         10.00 &   17.50 &  68.75 \\
    Chinese    &          0.00 &          0.00 &        0.00 &         6.25 &          10.00 &        13.75 &         15.00 &   11.25 &  67.50 \\
    German     &          0.00 &          0.00 &        0.00 &         5.00 &          10.00 &         6.25 &         10.00 &   10.00 &  72.50 \\
    Spanish    &          0.00 &          0.00 &        0.00 &         6.25 &          10.00 &        11.25 &         12.50 &   11.25 &  66.25 \\
    French     &          0.00 &          0.00 &        0.00 &         2.50 &           8.75 &        10.00 &         15.00 &   11.25 &  60.00 \\
    Italian    &          0.00 &          0.00 &        0.00 &         7.50 &          11.25 &        12.50 &         13.75 &   12.50 &  71.25 \\
    Portuguese &          0.00 &          0.00 &        0.00 &         6.25 &          10.00 &        10.00 &         13.75 &   12.50 &  63.75 \\
    Greek      &          0.00 &          0.00 &        0.00 &         3.75 &           8.75 &         7.50 &         11.25 &   10.00 &  61.25 \\
    Hungarian  &          0.00 &          0.00 &        0.00 &         5.00 &          11.25 &         8.75 &         11.25 &   11.25 &  65.00 \\
    Dutch      &          0.00 &          0.00 &        0.00 &         5.00 &          10.00 &        11.25 &         15.00 &   11.25 &  61.25 \\
    Finnish    &          0.00 &          0.00 &        0.00 &         5.00 &           7.50 &         8.75 &         10.00 &   10.00 &  65.00 \\
    Indonesian &          0.00 &          0.00 &        0.00 &         5.00 &          10.00 &         8.75 &         12.50 &    7.50 &  61.25 \\
    Turkish    &          0.00 &          0.00 &        0.00 &         8.75 &          10.00 &         7.50 &         10.00 &   10.00 &  51.25 \\
    Arabic     &          0.00 &          0.00 &        0.00 &         2.50 &           8.75 &        10.00 &          8.75 &   13.75 &  53.75 \\
    Vietnamese &          0.00 &          0.00 &        0.00 &         5.00 &          10.00 &        10.00 &         12.50 &   10.00 &  47.50 \\
    Bulgarian  &          0.00 &          0.00 &        0.00 &         6.25 &           7.50 &        12.50 &         10.00 &    8.75 &  68.75 \\
    Persian    &          0.00 &          0.00 &        0.00 &         2.50 &           8.75 &         7.50 &         11.25 &    8.75 &  62.50 \\
    Malay      &          0.00 &          0.00 &        0.00 &         3.75 &           7.50 &         6.25 &         11.25 &    8.75 &  63.75 \\
    Hebrew     &          0.00 &          0.00 &        0.00 &         2.50 &          10.00 &         8.75 &         10.00 &   18.75 &  60.00 \\
    Estonian   &          0.00 &          0.00 &        0.00 &         3.75 &           8.75 &        11.25 &         12.50 &   15.00 &  65.00 \\
    Tagalog    &          0.00 &          0.00 &        0.00 &         5.00 &           6.25 &         6.25 &         12.50 &    6.25 &  66.25 \\
    Afrikaans  &          0.00 &          0.00 &        0.00 &         5.00 &          11.25 &         8.75 &         13.75 &    8.75 &  68.75 \\
    \bottomrule
    \end{tabular}

    }

    \caption{Results (pass@1) of different models on Ruby across 23 natural languages.}
    \label{tab:appendix_ruby}
\end{table*}

\begin{table*}[]
    \centering
    \resizebox{0.9\textwidth}{!}{%
    \begin{tabular}{llllllllll}
    \toprule
    {} & CodeT5+(220M) & CodeT5+(770M) & CodeT5+(2B) & CodeGen2(1B) & CodeGen2(3.7B) & CodeGen2(7B) & CodeGen2(16B) & GPT-3.5 &  GPT-4 \\
    \midrule
    English    &          0.00 &          0.00 &        0.00 &         0.00 &           0.00 &         0.00 &          0.00 &    5.00 &  28.75 \\
    Russian    &          0.00 &          0.00 &        0.00 &         0.00 &           0.00 &         0.00 &          0.00 &   12.50 &  26.25 \\
    Chinese    &          0.00 &          0.00 &        0.00 &         0.00 &           0.00 &         0.00 &          0.00 &    2.50 &  18.75 \\
    German     &          0.00 &          0.00 &        0.00 &         0.00 &           0.00 &         0.00 &          0.00 &   10.00 &  26.25 \\
    Spanish    &          0.00 &          0.00 &        0.00 &         0.00 &           0.00 &         0.00 &          0.00 &    5.00 &  25.00 \\
    French     &          0.00 &          0.00 &        0.00 &         0.00 &           0.00 &         0.00 &          0.00 &    3.75 &  23.75 \\
    Italian    &          0.00 &          0.00 &        0.00 &         0.00 &           0.00 &         0.00 &          0.00 &    5.00 &  20.00 \\
    Portuguese &          0.00 &          0.00 &        0.00 &         0.00 &           0.00 &         0.00 &          0.00 &    8.75 &  15.00 \\
    Greek      &          0.00 &          0.00 &        0.00 &         0.00 &           0.00 &         0.00 &          0.00 &    8.75 &  18.75 \\
    Hungarian  &          0.00 &          0.00 &        0.00 &         0.00 &           0.00 &         0.00 &          0.00 &    8.75 &  32.50 \\
    Dutch      &          0.00 &          0.00 &        0.00 &         0.00 &           0.00 &         0.00 &          0.00 &   10.00 &  23.75 \\
    Finnish    &          0.00 &          0.00 &        0.00 &         0.00 &           0.00 &         0.00 &          0.00 &    8.75 &  28.75 \\
    Indonesian &          0.00 &          0.00 &        0.00 &         0.00 &           0.00 &         0.00 &          0.00 &    5.00 &  21.25 \\
    Turkish    &          0.00 &          0.00 &        0.00 &         0.00 &           0.00 &         0.00 &          0.00 &    6.25 &  41.25 \\
    Arabic     &          0.00 &          0.00 &        0.00 &         0.00 &           0.00 &         0.00 &          0.00 &    7.50 &  10.00 \\
    Vietnamese &          0.00 &          0.00 &        0.00 &         0.00 &           0.00 &         0.00 &          0.00 &    2.50 &  27.50 \\
    Bulgarian  &          0.00 &          0.00 &        0.00 &         0.00 &           0.00 &         0.00 &          0.00 &   13.75 &  25.00 \\
    Persian    &          0.00 &          0.00 &        0.00 &         0.00 &           0.00 &         0.00 &          0.00 &   15.00 &  27.50 \\
    Malay      &          0.00 &          0.00 &        0.00 &         0.00 &           0.00 &         0.00 &          0.00 &    2.50 &  22.50 \\
    Hebrew     &          0.00 &          0.00 &        0.00 &         0.00 &           0.00 &         0.00 &          0.00 &   11.25 &  17.50 \\
    Estonian   &          0.00 &          0.00 &        0.00 &         0.00 &           0.00 &         0.00 &          0.00 &    7.50 &  28.75 \\
    Tagalog    &          0.00 &          0.00 &        0.00 &         0.00 &           0.00 &         0.00 &          0.00 &    6.25 &  21.25 \\
    Afrikaans  &          0.00 &          0.00 &        0.00 &         0.00 &           0.00 &         0.00 &          0.00 &    7.50 &  20.00 \\
    \bottomrule
    \end{tabular}

    }

    \caption{Results (pass@1) of different models on Scala across 23 natural languages.}
    \label{tab:appendix_scala}
\end{table*}

\begin{table*}[]
    \centering
    \resizebox{0.9\textwidth}{!}{%
    \begin{tabular}{llllllllll}
    \toprule
    {} & CodeT5+(220M) & CodeT5+(770M) & CodeT5+(2B) & CodeGen2(1B) & CodeGen2(3.7B) & CodeGen2(7B) & CodeGen2(16B) & GPT-3.5 &  GPT-4 \\
    \midrule
    English    &          0.00 &          0.00 &        0.00 &         3.75 &           6.25 &         8.75 &         12.50 &    3.75 &  52.50 \\
    Russian    &          0.00 &          1.25 &        0.00 &         2.50 &           5.00 &         7.50 &         12.50 &    3.75 &  58.75 \\
    Chinese    &          0.00 &          1.25 &        0.00 &         3.75 &           5.00 &         7.50 &         10.00 &    1.25 &  52.50 \\
    German     &          0.00 &          0.00 &        0.00 &         5.00 &           3.75 &         5.00 &          8.75 &    3.75 &  55.00 \\
    Spanish    &          0.00 &          0.00 &        0.00 &         3.75 &           2.50 &         6.25 &         11.25 &    2.50 &  55.00 \\
    French     &          0.00 &          0.00 &        0.00 &         2.50 &           5.00 &         3.75 &          7.50 &    0.00 &  53.75 \\
    Italian    &          0.00 &          0.00 &        0.00 &         2.50 &           3.75 &         5.00 &         12.50 &    1.25 &  56.25 \\
    Portuguese &          0.00 &          0.00 &        0.00 &         3.75 &           3.75 &         7.50 &          8.75 &    1.25 &  53.75 \\
    Greek      &          0.00 &          1.25 &        0.00 &         3.75 &           5.00 &         6.25 &          8.75 &    2.50 &  51.25 \\
    Hungarian  &          0.00 &          0.00 &        0.00 &         3.75 &           6.25 &         6.25 &          7.50 &    2.50 &  57.50 \\
    Dutch      &          0.00 &          0.00 &        0.00 &         3.75 &           5.00 &         6.25 &         10.00 &    6.25 &  56.25 \\
    Finnish    &          0.00 &          0.00 &        0.00 &         3.75 &           6.25 &         6.25 &          7.50 &    0.00 &  55.00 \\
    Indonesian &          0.00 &          0.00 &        0.00 &         3.75 &           5.00 &         7.50 &         11.25 &    0.00 &  58.75 \\
    Turkish    &          0.00 &          0.00 &        0.00 &         2.50 &           5.00 &         5.00 &          7.50 &    1.25 &  57.50 \\
    Arabic     &          0.00 &          1.25 &        0.00 &         2.50 &           3.75 &         5.00 &          6.25 &    2.50 &  55.00 \\
    Vietnamese &          0.00 &          0.00 &        0.00 &         2.50 &           3.75 &         6.25 &         10.00 &    2.50 &  56.25 \\
    Bulgarian  &          0.00 &          1.25 &        0.00 &         2.50 &           2.50 &         6.25 &         11.25 &    1.25 &  57.50 \\
    Persian    &          0.00 &          1.25 &        0.00 &         2.50 &           5.00 &         6.25 &          8.75 &    1.25 &  58.75 \\
    Malay      &          0.00 &          0.00 &        0.00 &         3.75 &           6.25 &         6.25 &         10.00 &    1.25 &  55.00 \\
    Hebrew     &          0.00 &          1.25 &        0.00 &         3.75 &           3.75 &         5.00 &         10.00 &    5.00 &  55.00 \\
    Estonian   &          0.00 &          0.00 &        0.00 &         2.50 &           5.00 &         6.25 &          6.25 &    2.50 &  56.25 \\
    Tagalog    &          0.00 &          0.00 &        0.00 &         3.75 &           6.25 &         5.00 &          6.25 &    2.50 &  56.25 \\
    Afrikaans  &          0.00 &          0.00 &        0.00 &         3.75 &           3.75 &         5.00 &          6.25 &    5.00 &  57.50 \\
    \bottomrule
    \end{tabular}

    }

    \caption{Results (pass@1) of different models on Swift across 23 natural languages.}
    \label{tab:appendix_swift}
\end{table*}

\begin{table*}[]
    \centering
    \resizebox{0.9\textwidth}{!}{%
    \begin{tabular}{llllllllll}
    \toprule
    {} & CodeT5+(220M) & CodeT5+(770M) & CodeT5+(2B) & CodeGen2(1B) & CodeGen2(3.7B) & CodeGen2(7B) & CodeGen2(16B) & GPT-3.5 &  GPT-4 \\
    \midrule
    English    &          0.00 &          1.25 &        1.25 &         7.50 &          16.25 &        21.25 &         22.50 &   11.25 &  75.00 \\
    Russian    &          0.00 &          2.50 &        0.00 &         8.75 &          11.25 &        18.75 &         23.75 &    8.75 &  70.00 \\
    Chinese    &          0.00 &          1.25 &        0.00 &         6.25 &          13.75 &        18.75 &         23.75 &    3.75 &  68.75 \\
    German     &          0.00 &          1.25 &        0.00 &        10.00 &          13.75 &        20.00 &         20.00 &    3.75 &  70.00 \\
    Spanish    &          0.00 &          1.25 &        0.00 &        10.00 &          11.25 &        23.75 &         21.25 &    5.00 &  73.75 \\
    French     &          0.00 &          0.00 &        1.25 &         8.75 &          13.75 &        20.00 &         21.25 &    1.25 &  71.25 \\
    Italian    &          0.00 &          1.25 &        0.00 &         8.75 &          13.75 &        16.25 &         20.00 &    5.00 &  72.50 \\
    Portuguese &          0.00 &          0.00 &        0.00 &         7.50 &          13.75 &        20.00 &         21.25 &    6.25 &  71.25 \\
    Greek      &          0.00 &          3.75 &        0.00 &        10.00 &          12.50 &        15.00 &         20.00 &    5.00 &  65.00 \\
    Hungarian  &          0.00 &          0.00 &        0.00 &        10.00 &          12.50 &        13.75 &         18.75 &    5.00 &  76.25 \\
    Dutch      &          0.00 &          0.00 &        0.00 &         7.50 &          16.25 &        20.00 &         20.00 &    6.25 &  70.00 \\
    Finnish    &          0.00 &          0.00 &        0.00 &        10.00 &          15.00 &        16.25 &         20.00 &    6.25 &  73.75 \\
    Indonesian &          0.00 &          0.00 &        0.00 &         6.25 &          15.00 &        18.75 &         20.00 &    2.50 &  71.25 \\
    Turkish    &          0.00 &          2.50 &        0.00 &        11.25 &          12.50 &        17.50 &         18.75 &    2.50 &  68.75 \\
    Arabic     &          0.00 &          3.75 &        0.00 &        10.00 &          12.50 &        17.50 &         25.00 &    3.75 &  66.25 \\
    Vietnamese &          0.00 &          1.25 &        0.00 &        11.25 &          15.00 &        15.00 &         22.50 &    6.25 &  70.00 \\
    Bulgarian  &          0.00 &          2.50 &        0.00 &        10.00 &          11.25 &        17.50 &         21.25 &    7.50 &  70.00 \\
    Persian    &          0.00 &          5.00 &        0.00 &         8.75 &          15.00 &        22.50 &         18.75 &    5.00 &  72.50 \\
    Malay      &          0.00 &          1.25 &        0.00 &         8.75 &          12.50 &        20.00 &         18.75 &    6.25 &  71.25 \\
    Hebrew     &          0.00 &          2.50 &        0.00 &         8.75 &          12.50 &        17.50 &         20.00 &    5.00 &  71.25 \\
    Estonian   &          0.00 &          0.00 &        0.00 &        11.25 &          12.50 &        16.25 &         22.50 &    7.50 &  72.50 \\
    Tagalog    &          0.00 &          0.00 &        0.00 &         8.75 &          12.50 &        17.50 &         18.75 &    6.25 &  72.50 \\
    Afrikaans  &          0.00 &          0.00 &        0.00 &        11.25 &          13.75 &        17.50 &         21.25 &    1.25 &  70.00 \\
    \bottomrule
    \end{tabular}

    }

    \caption{Results (pass@1) of different models on TypeScript across 23 natural languages.}
    \label{tab:appendix_typescript}
\end{table*}

\end{document}